\documentclass[lettersize,journal]{IEEEtran}
\usepackage{amsmath,amsfonts}
\usepackage{algorithmic}
\usepackage{algorithm}
\usepackage{array}
\usepackage{textcomp}
\usepackage{stfloats}
\usepackage{url}
\usepackage{verbatim}
\usepackage{graphicx}
\usepackage{cite}
\usepackage{bm}
\usepackage{subcaption}
\allowdisplaybreaks[4]
\usepackage{relsize}
\usepackage{xcolor}
\begin{document}

\title{\huge
Learning to Optimize Edge Robotics: A Fast Integrated Perception-Motion-Communication Approach
\vspace{-0.1in}
}
\author{
Dan Guo, Xibin Jin, Shuai Wang$^{\dag}$, Zhigang Wen$^{\dag}$, Miaowen Wen, and  Chengzhong Xu,~\emph{Fellow, IEEE}
\vspace{-0.2in}
\thanks{
This work was supported by the National Natural Science Foundation of China (Grant No. 62371444).
Dan Guo and Zhigang Wen are with Beijing University of Posts and Telecommunications, Beijing, China. 
Xibin Jin and Miaowen Wen are with the South China University of Technology, Guangzhou, China.
Shuai Wang is with the Shenzhen Institutes of Advanced Technology, Chinese Academy of Sciences, Shenzhen, China. 
Chengzhong Xu is with the University of Macau, Macau, China.

Corresponding author: Shuai Wang ({\tt\footnotesize s.wang@siat.ac.cn}) and Zhigang Wen ({\tt\footnotesize zwen@bupt.edu.cn}). 
}
}
\maketitle

\begin{abstract}
Edge robotics involves frequent exchanges of large-volume multi-modal data. 
Existing methods ignore the interdependency between robotic functionalities and communication conditions, leading to excessive communication overhead. 
This paper revolutionizes edge robotics systems through integrated perception, motion, and communication (IPMC). As such, robots can dynamically adapt their communication strategies (i.e., compression ratio, transmission frequency, transmit power) by leveraging the knowledge of robotic perception and motion dynamics, thus reducing the need for excessive sensor data uploads. Furthermore, by leveraging the learning to optimize (LTO) paradigm, an imitation learning neural network is designed and implemented, which reduces the computational complexity by over 10x compared to state-of-the art optimization solvers. Experiments demonstrate the superiority of the proposed IPMC and the real-time execution capability of LTO.
\end{abstract}

\vspace{-0.1in}
\section{Introduction}
Edge robotics (ER) enables resource-constrained mobile robots to offload computation-intensive tasks to edge servers\cite{CE-IoT, edge_robot,edge_new,edge_ai, rluo} . 
Its challenge lies in transmitting massive multi-modal data streams under limited robot power. 
Classical strategies adopt communication throughput \cite{sumrate_maximization} or user fairness \cite{maxmin_fairness} as their design objectives. 
Nonetheless, these metrics are not tailored for ER and cannot account for the 
data diversities across different modalities and frames.  

Recently, multi-modal communication approaches have been proposed to address the above issue, by transmitting compressed sensor data to edge server and dynamically adjusting the compression ratio \cite{STS, Optimizing_Data_Transmission, Comparison_of_Edge_Computing, Green_sEdge,incre_server}.
For instance, a self-adaptive transmission scheme (STS) \cite{STS} was proposed to deal with the real-time video transmission under a changing environment of network. 
Moreover, actor-critic learning methods were developed to map environmental variables to optimal operations \cite{Real-time,luo2024deep}.
However, all these methods ignore the inter-dependency between robotic functionalities and communication
conditions. For instance, no communication is required at all if the robot involves no movement (i.e., a sequence of data contains the same scenes).
In addition, they often overlook cross-modal resource allocation.
As such, they may lead to sensor data distortion or excessive communication overhead.

This paper revolutionizes ER systems through integrated perception, motion, and communication (IPMC). 
Our proposed IPMC dynamically adjusts communication strategies based on perception and motion dynamics, prioritizing data towards scenarios with higher variability, so as to facilitate data recovery and enhance data fidelity at the receiver. 
Specifically, we formulate a noncovonex IPMC problem that jointly optimizes robot compression ratios, down-sampling rates, and transmit powers.
To proceed, we transform the nonconvex IPMC into an equivalent geometric programming (GP) form that can be optimally solved by interior point method.
But since solving GP is time-consuming, we further leverage the learning-to-optimize (LTO) framework to imitate the solving procedure of GP using deep neural network (DNN).
Experimental results demonstrate the superiority of IPMC over benchmark methods in various scenarios. 
Moreover, the proposed LTO-IPMC reduces the computation time by over $10$x compared to GP-IPMC, while achieving a close-to-GP performance.

The remainder of this paper is organized as follows.
Section \ref{section2} describes system model. 
Section \ref{section3} proposes our models and formulate the IPMC problem.
Subsequently, Section~\ref{section4} presents the GP and LTO algorithms. 
Section~\ref{section5} presents experimental results. 
Finally, Section~\ref{section6} concludes this work.

\begin{figure*}[!t]
\centering
    \begin{subfigure}[b]{0.26\textwidth}
        \includegraphics[width=\linewidth]{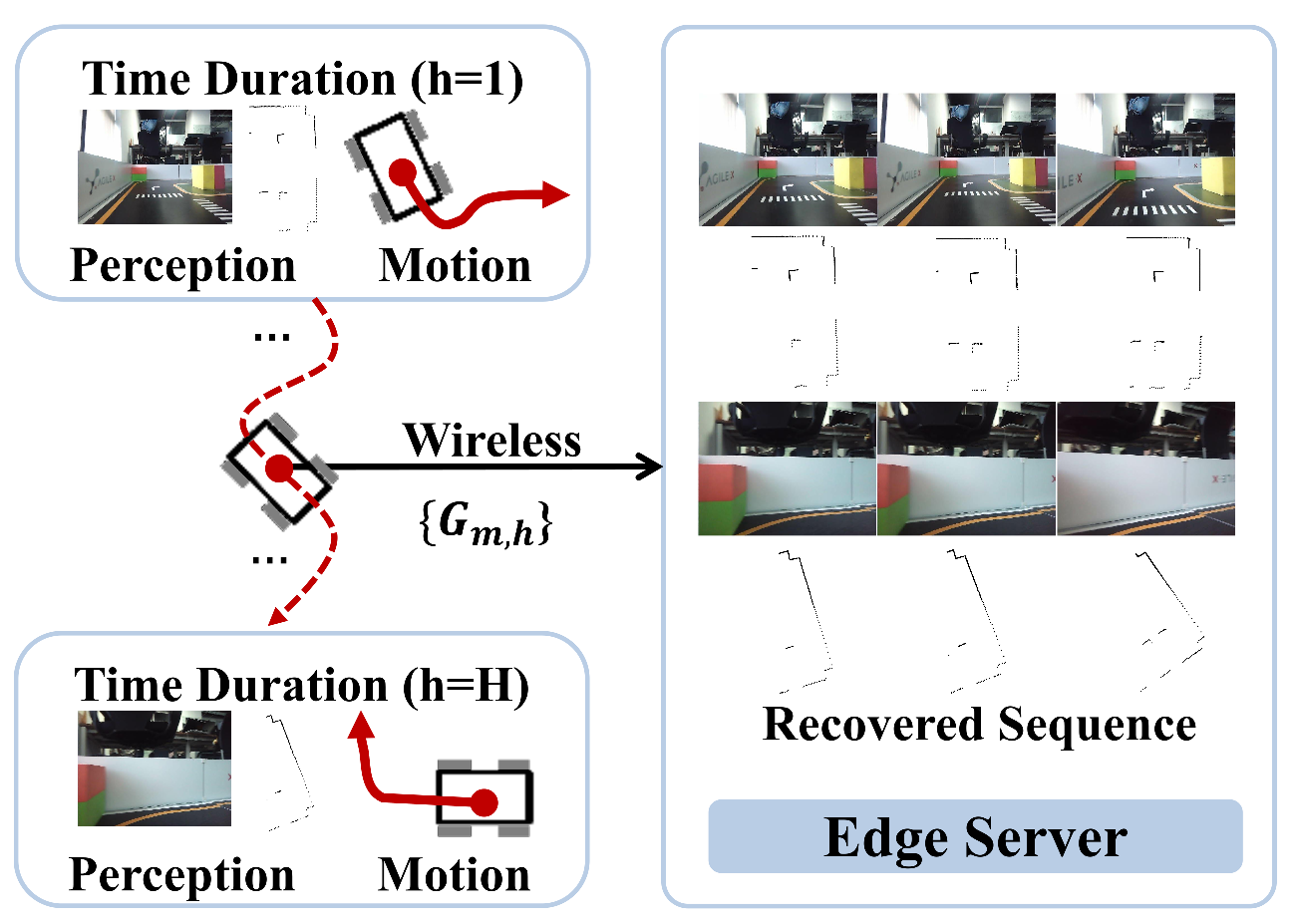}
        \caption{{ER system model.}}
        \label{fig:1a}
    \end{subfigure}
    \begin{subfigure}[b]{0.38\textwidth}
        \includegraphics[width=\linewidth]{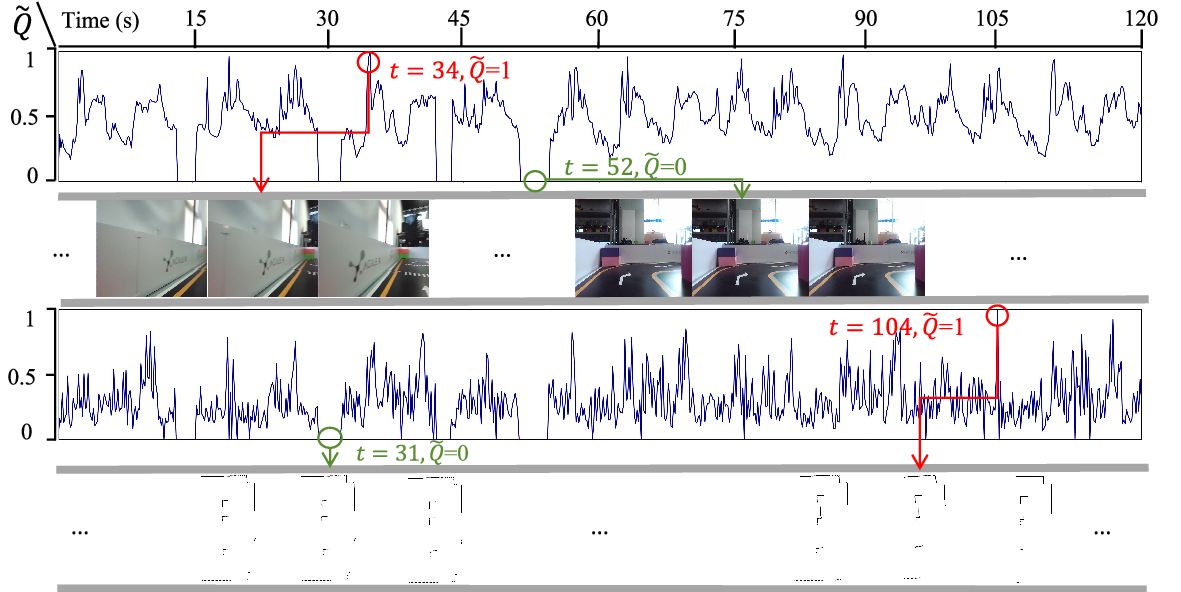}
        \caption{{Perception-aware objective.}}
        \label{fig:1b}
    \end{subfigure}
        \begin{subfigure}[b]{0.30\textwidth}
        \includegraphics[width=\linewidth]{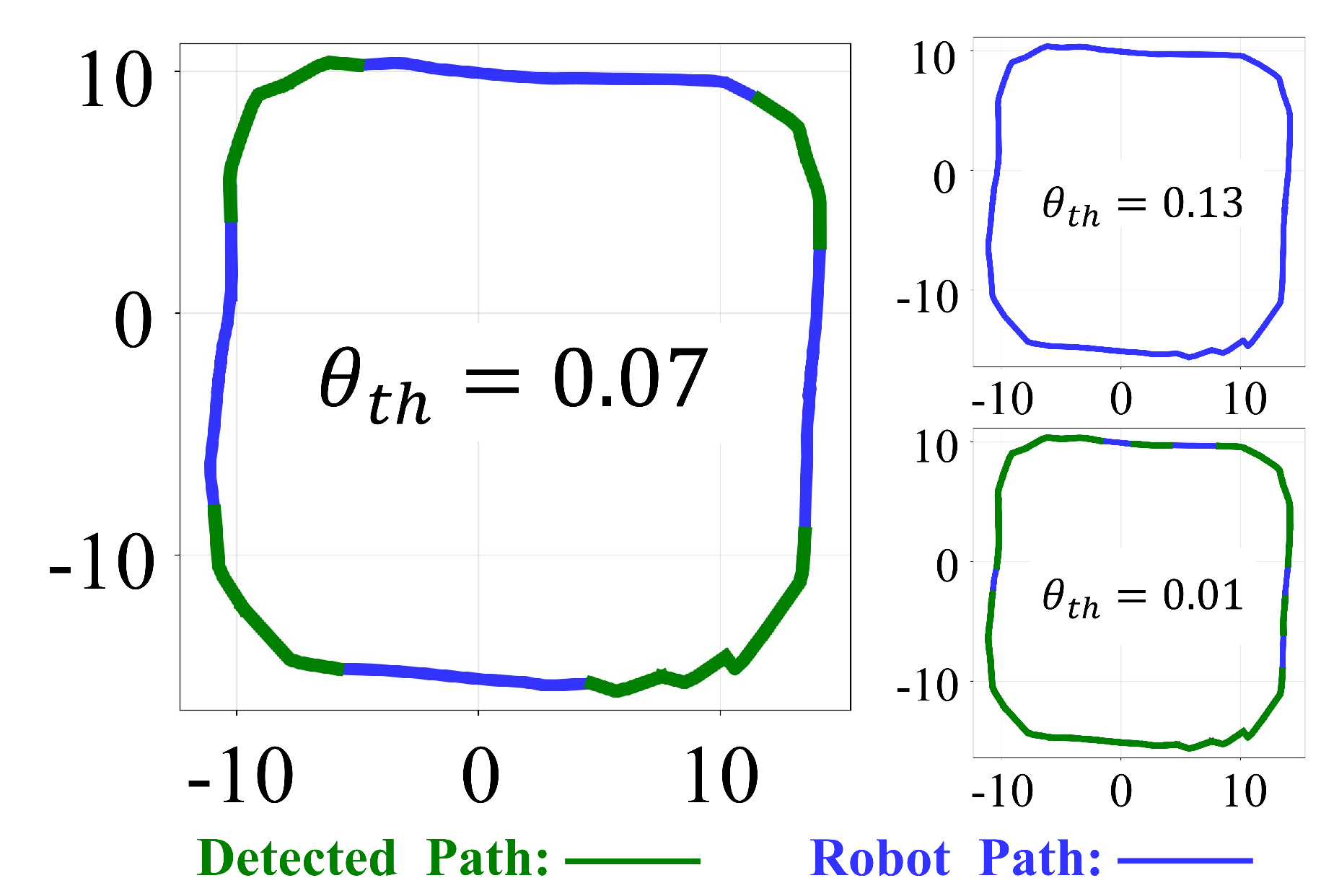}
        \caption{{Motion-aware constraint.}}
        \label{fig:1c}
    \end{subfigure}
    \vspace{-0.05in} 
    \caption{{System model of edge robotics and proposed perception-motion-communication models.}}
    \label{fig:fig1}
    \vspace{-0.2in} 
\end{figure*}

\vspace{-0.1in}
\section{System Model}\label{section2}

We consider an ER system shown in Fig.~\ref{fig:fig1}a, which consists of a mobile robot and an edge server. The task is to upload multi-modal sensory data from robot to server over wireless communications.
The robot states and measurements are viewed as discrete measurements in the time domain.
The entire task duration is divided into $H$ sub-durations, and each sub-duration has $L$ data frames, where the period between two consecutive states is denoted as $\Delta t$. 
Data frames in each sub-duration share the same transmission profile. 

At the $l$-th time slot ($l \in \mathcal{L}$ with $\mathcal{L}=\{1,\cdots,L\}$) of the $h$-th sub-duration ($h \in \mathcal{H}$ with $\mathcal{H}=\{1,\dots,H\}$), the robot state is $\mathbf{s}_{h,l}=(a_{h,l},e_{h,l},\omega_{h,l})$, where $\mathbf{y}_{h,l}=(a_{h,l},e_{h,l})$ and $\omega_{h,l}$ are associated position and orientation, respectively. 
The state evolution is
\begin{align}
    {\mathbf{s}_{h,l+1}} = {\mathbf{s}_{h,l}} + f({\mathbf{s}_{h,l}},{\mathbf{u}_{h,l}})\Delta t, \ \forall l\in\mathcal{L},
    \label{eq:dynamics}
\end{align}
where $\mathbf{u}_{h,l}=(v_{h,l},\psi_{h,l})$ is the robot action, with $v_{h,l}$ and $\psi_{h,l}$ being the linear velocity and steering angle, respectively. The Ackerman kinematic model $f$ is \cite{ackermanwheel}:
\begin{align}
f({\mathbf{s}_{h,l}},{\mathbf{u}_{h,l}}) = \left[
    {{v_{h,l}}\cos ({\omega _{h,l}})}, 
    {{v_{h,l}}\sin ({\omega _{h,l}})}, 
    {\frac{{{v_{h,l}}\tan {\psi _{h,l}}}}{W}}
  \right]^T. \nonumber
  \end{align}
where constant $W$ denotes the distance between the front and rear axles, which is set to $W=0.3$\,m in our experiment.

The robot measurements involve $M$ modalities (e.g., camera, lidar).
Let $\mathbf{x}_{m,h,l}\in {\mathbb{R}^{N_m \times 1}}$ denote sensor data of the $m$-th modality. 
The average data volume of each sample for modality $m$ is $Z_{m}$.
To transmit samples with low overheads, the down-sampling rate and the compression ratio of the $m$-th modality at the $h$-th sub-duration are $d_{m,h}\in[0,1]$ and $c_{m,h}\in[0,1]$, respectively.
To ensure data transmission, we must have $d_{m,h}\in[d_{\mathrm{min}},1]$ and $c_{m,h}\in[c_{\mathrm{min}},1]$, where $d_{\mathrm{min}}$ and $c_{\mathrm{min}}$ are minimum values of down-sampling and compression ratios, respectively. 

We consider orthogonal frequency division multiplexing with a total bandwidth of $B$, and the $m$-th modality possesses adjacent subcarriers with a bandwidth $B_m$ ($\sum B_m=B$).
Under such a scheme, the robot data-rate needs to satisfy 
\begin{align}\label{rate}
    &\Delta t B_m\log _{2}\left(1+\frac{G_{m,h}p_{m,h}}{\sigma^2}\right) 
    \geq d_{m,h}c_{m,h}Z_{m},
\end{align}
where $p_{m,h}$ is the transmit power of modality $m$\footnote{Without loss of generality, we consider modality-wise instead of subcarrier-wise power control to reduce the design complexity.}, $\sigma^2$ is the power of additive white Gaussian noise (AWGN), and $G_{m,h}$ is the channel gain from the robot to the server. 
The average power is subject to a budget $P_{\rm{sum}}$, i.e., 
\begin{align}\label{power}
&\frac{1}{H}\sum\limits_{m=1}^M\sum\limits_{h=1}^H p_{m,h}\leq P_{\rm{sum}}.
\end{align}
After receiving the robot data, the edge server adopts de-compression to enhance the data quality and interpolation to recover the missing data frames.

\section{Integrated Perception, Motion, and Communication for Edge Robotics}\label{section3}

In the considered ER system, the controllable variables are down-sampling ratios $\mathbf{D}=[d_{1,1},\cdots,d_{M,H}]^T$, compression ratios $\mathbf{C}=[c_{1,1},\cdots,c_{M,H}]^T$, and transmit powers $\mathbf{P}=[p_{1,1},\cdots,p_{M,H}]^T$. 
The problem of ER is how to support the uploading of sensor data $\{\mathbf{x}_{m,h,l}\}$ under stringent communication budget $P_{\mathrm{sum}}$ by jointly controlling $(\mathbf{D},\mathbf{C},\mathbf{P})$.

We present an IPMC solution, which addresses the above problem through dynamic reshaping of $(\mathbf{D},\mathbf{C},\mathbf{P})$ according to both scenario change rates and robot motion conditions. 
As such, we can prioritize resources towards more dynamic situations by reducing powers in static situations.

\vspace{-0.1in}
\subsection{Perception-Aware Objective Function}

Let $\widehat{\textbf{x}}_{m,h,l}$ denote the server recovered data, which is a function of $(\mathbf{D},\mathbf{C})$.
The average data distortion $\mathsf{MSE}_{m,h}$ is
\begin{align}
    \mathsf{MSE}_{m,h}(\mathbf{D},\mathbf{C})=\frac{1}{L}\sum\limits_{l=1}^L  
    {\frac{1}{N_{m}}\|\textbf{x}_{m,h,l}-\widehat{\textbf{x}}_{m,h,l}}\|^{2}.
    \label{eq:distortion}
\end{align}
Our goal is to minimize the worst distortion across all sub-durations and modalities, i.e., 
\begin{align}\label{MSE}
\min~\max_{m,h}~\mathsf{MSE}_{m,h}(\mathbf{D},\mathbf{C}).
\end{align}
The major obstacle to solving \eqref{MSE} is that the cost function is unknown prior to data transmission due to the existence of $\widehat{\textbf{x}}_{m,h,l}$.
Here we propose a surrogate function optimization approach. 
First, we introduce the so-called scenario change indicator (SCI) $\mathbf{Q}=[Q_{1,1},\cdots,Q_{1,H};\cdots;Q_{M,1},\cdots, Q_{M,H}] \in {\mathbb{R}^{M \times H}}$, to measure the variation of multi-modal data\cite{mse_change}: 
\begin{align}\label{Qmh}
    Q_{m,h}=\frac{1}{L}\sum\limits_{l=1}^L  
    {\frac{1}{N_{m}}\|\textbf{x}_{m,h,l}-\textbf{x}_{m,h,l-1}}\|^{2}.
\end{align}
By comparing equations \eqref{eq:distortion} and \eqref{Qmh}, it is found that $Q_{m,h}$ is a related function of $\mathsf{MSE}_{m,h}$ that replaces the unknown $\widehat{\textbf{x}}_{m,h,l}$ with a known $\textbf{x}_{m,h,l-1}$. 
Such differential approximation is tight if the current missing frame $\widehat{\textbf{x}}_{m,h,l}$ is exactly recovered from the previous frame.  

Since different data modalities may have different magnitudes, we further normalize $Q_{m,h}$ into  
\begin{align}
    \widetilde{Q}_{m,h}=\frac{Q_{m,h}-\min(\mathbf{Q_{m}})}{\max(\mathbf{Q_{m}})-\min(\mathbf{Q_{m}})},
    \label{eq:Q}
\end{align}
where $\mathbf{Q}_m=[Q_{m,1}, Q_{m,2}, \dots, Q_{m,H}]$ is the SCI for modality $m$ across all sub-durations, and functions $\min(\mathbf{Q}_{m})$ and $\max(\mathbf{Q}_{m})$ compute the minimum and maximum values of $\mathbf{Q}_{m}$, respectively. With $\widetilde{Q}_{m,h}$, we can define the relative SCI $\widetilde{\mathbf{Q}}$ accordingly. 
Based on relative SCIs, we propose to optimize a surrogate function 
$\min \, C_0(\mathbf{D},\mathbf{C})$, where 
\begin{align}
 C_0(\mathbf{D},\mathbf{C}):=\max_{m,h}~\tilde{Q}_{m,h}^\alpha{d_{m,h}^{-1}c_{m,h}^{-1}},
    \label{eq:efficiency}
\end{align}
where $\alpha$ is a tuning parameter, and the weighting factors $\{d_{m,h},c_{m,h}\}$ are added to denominator since the distortion is inversely proportional to $\{d_{m,h},c_{m,h}\}$.

To evaluate the proposed perception-aware objective, we compare the $\widetilde{Q}_{m,h}$ values and the experimental data in Fig.~\ref{fig:fig1}b. 
Our model matches the real data very well for both camera and lidar data. 
Particularly, at time step $t=34$, the robot steers to the right, and the image view changes quickly. In such a case, we have $\widetilde{Q}_{1,h}=1$. 
In contrast, at time step $t=52$, the robot stops and the image view remains the same. In this case, we have $\widetilde{Q}_{1,h}=0$. 
This example clearly demonstrates the accuracy of our model for image modality.
The model also fits the experimental data for lidar very well, but the peak value of $\widetilde{Q}_{m,h}$ is different for different $m$.

\vspace{-0.6\baselineskip} 

\subsection{Motion-Dynamics Aware Communication Constraints}

In practice, the robot sensor data may involve high uncertainties when the robot steers.
To ensure fidelity, it is necessary to avoid down-sampling and compression at turning points.  
To predict the steering angle, we propose to adopt forward induction based on robot state transition \eqref{eq:dynamics} at each sub-duration. 
In particular, 
given the robot state $\mathbf{s}_{h,l}$ at the time slot $l$ in the sub-duration $h$, the future $L-1$ states can be estimated using model predictive control (MPC):
\begin{subequations} 
\begin{align}
\mathsf{MPC}: 
&\min_{\{\mathbf{s}_{h,l}, \mathbf{u}_{h,l}\}_{l=1}^{L}} \sum^{L}_{l=1} 
\left\|\mathbf{s}_{h,l}-\mathbf{s}_{h,l}^\diamond\right\|^2 \\
&\text{s.t.} \quad \mathbf{s}_{h,l+1} = \mathbf{s}_{h,l} + f(\mathbf{s}_{h,l},\mathbf{u}_{h,l})\Delta t,~\forall l \\
&\qquad \mathbf{u}_{\min} \preceq \mathbf{u}_{h,l} \preceq \mathbf{u}_{\max},~\forall l, \\
&\qquad \bm{\beta}_{\min} \preceq \mathbf{u}_{h,l+1} - \mathbf{u}_{h,l} \preceq \bm{\beta}_{\max},~\forall l,
\end{align}
\end{subequations}
where $\{\mathbf{s}_{h,l}^\diamond\}$ are the target waypoints (which are predetermined), ${\mathbf{u}_{\min }}$ and ${\mathbf{u}_{\max }}$ are the minimum and maximum values of the control vector, respectively, and $\bm{{\beta}}_{\min}$ and ${\bm{\beta}}_{\max}$ are the associated minimum and maximum accelerations, respectively.
The above problem can be solved by off-the-shelf solver (e.g., cvxpy) in an offline manner. 

Let $\{\mathbf{s}_{h,l}^*\}_{l=1}^{L}$ with $\mathbf{s}_{h,l}^*=(a_{h,l}^*,e_{h,l}^*,w_{h,l}^*)$ denote the optimal solution to problem $\mathsf{MPC}$. 
We calculate the trajectory curvature $\theta_h$ during time slot $h$ based on $\{\mathbf{s}_{h,l}^*\}$ using \cite[Eqn. 1]{curvature}.
With $\theta_h$, we introduce the motion-aware communication constraints 
\begin{subequations} 
\begin{align}
&d_{m,h} \geq \psi_h\Delta d + d_{\mathrm{min}}, \quad  \hfill \forall m,h, \label{dmh}
\\
&c_{m,h} \geq 
\psi_h\Delta c
+ c_{\mathrm{min}}, \quad  \hfill \forall m,h,  \label{cmh}
\end{align}
\end{subequations}
where $\bm{\psi}=[\mathbb{I}_{\{\theta_1\geq \theta_{\mathrm{th}}\}}(\theta_1),\cdots,\mathbb{I}_{\{\theta_H\geq \theta_{\mathrm{th}}\}}(\theta_H)]^T \in {\mathbb{R}^{H \times 1}}$ is a motion-aware vector, and $(\Delta d,\Delta c)$ are additional increments added up to the lower bounds. 
The indicator function $\mathbb{I}_{\mathcal{A}}(a)=1$ if $a\in \mathcal{A}$ and $\mathbb{I}_{\mathcal{A}}(a)=0$ otherwise. 

{
To verify the proposed motion-aware constraint, we compare the detected path and the entire robot path in Fig.~\ref{fig:fig1}c. 
With a curve threshold of $\theta_{\mathrm{th}} = 0.07$, all turning paths are correctly detected. Note that improper $\theta_{\mathrm{th}}$ can misclassify straight parts as turning paths (i.e., $\theta_{\mathrm{th}} = 0.01$) or misclassify all parts as straight paths (i.e., $\theta_{\mathrm{th}}=0.13$).
}

\subsection{IPMC Problem Formulation}

Based on the results in Section III-A and III-B, the scenario-aware communication problem is formulated as 
\begin{subequations}
\begin{align}
    \mathsf{P}: \min\limits_{\mathbf{D},\mathbf{C},\mathbf{P}}~
    &C_0(\mathbf{D},\mathbf{C})
    \label{eq:objective}
    \ \\
    \text { s.t. }~~&\mathrm{constraints} \ \eqref{rate}, \eqref{power}, \eqref{dmh}, \eqref{cmh}.
\end{align}
\end{subequations}
Problem $\mathsf{P}$ is nontrivial to solve due to two reasons. 
First, $\mathsf{P}$ is a non-convex problem due to the nonlinear coupling of variables $d_{m,h}$ and $c_{m,h}$ in \eqref{eq:objective} and \eqref{rate}.
Second, the dimensionality of $\mathbf{D},\mathbf{C},\mathbf{P}$ may be large, leading to time-consuming problem solving procedure.

\vspace{-0.1in}
\section{Learning-to-Optimize Algorithm}\label{section4}

This section will tackle the first challenge
by transforming $\mathsf{P}$ into a GP form, and tackle the second challenge by imitating the GP algorithm using a DNN, so as to accelerate the mapping from scenario inputs $\{\mathbf{Z},\widetilde{\mathbf{Q}},\bm{\psi}\}$ to outputs $\{\mathbf{D}^{*},\mathbf{C}^{*},\mathbf{P}^{*}\}$.

\vspace{-0.1in}
\subsection{Geometric Programming}

The GP algorithm converts $\mathsf{P}$ by leveraging a set of variable transformations. 
Specifically, consider the following changes of variables: 
$\widehat{d}_{m,h}=\ln d_{m,h}$ and $\widehat{c}_{m,h}=\ln c_{m,h}, \ \forall m,h$. 
Accordingly, the objective function \eqref{eq:objective} in $\mathsf{P}$ is equivalently rewritten as 
\begin{align}
C_0(\mathbf{D},\mathbf{C})&=\max_{m,h}~\tilde{Q}_{m,h}^\alpha d_{m,h}^{-1}c_{m,h}^{-1}
\nonumber\\
&=
\max_{m,h}~
\mathrm{exp}\left[\ln (\tilde{Q}_{m,h}^\alpha)-\widehat{d}_{m,h}-\widehat{c}_{m,h}\right].
\end{align}
By further leveraging the monotonicity property of exponential function $\mathrm{exp}(\cdot)$, we can define 
\begin{align}
C_1(\widehat{\mathbf{D}},\widehat{\mathbf{C}})
=
\max_{m,h}~\left[\ln (\tilde{Q}_{m,h}^\alpha)-\widehat{d}_{m,h}-\widehat{c}_{m,h}\right],
\end{align}
and it is clear that minimizing $C_0(\mathbf{D},\mathbf{C})$ is equivalent to minimizing 
$C_1(\widehat{\mathbf{D}},\widehat{\mathbf{C}})$.
On the other hand, constraint \eqref{rate} in $\mathsf{P}$ is equivalently rewritten as
\begin{align}
&\Delta t B_m \log _{2}\left(1+\frac{G_{m,h}p_{m,h}}{\sigma^2}\right) 
    \nonumber\\
    &\geq \,\mathrm{exp}\left(\ln Z_{m,h}+\widehat{d}_{m,h}+\widehat{c}_{m,h}\right), \ \forall m,h. 
    \label{reformulated_1}
\end{align}

With the above transformations, the primal problem $\mathsf{P}$ is equivalently transformed into 
\begin{subequations}
\begin{align}
    \mathsf{P}_1: \min\limits_{\widehat{\mathbf{D}},\widehat{\mathbf{C}},\mathbf{P}}~&C_1(\widehat{\mathbf{D}},\widehat{\mathbf{C}})
     \\
    \text { s.t. }~~&\mathrm{constraints}~\eqref{reformulated_1}, \eqref{power},
    \\
    &\widehat{d}_{m,h} \geq \ln \left(\psi_h\Delta d + d_{\mathrm{min}}\right),~\forall m,h,
    \\
    &\widehat{c}_{m,h} \geq 
    \ln \left(
    \psi_h\Delta c
    + c_{\mathrm{min}}
    \right), ~\forall m,h.
\end{align}
\end{subequations}
Now problem $\mathsf{P}_1$ becomes a convex optimization problem, which can be optimally solved using off-the-shelf software (e.g., cvxpy). 
Denoting the optimal solution to $\mathsf{P}_1$ as $\{\widehat{d}^{*}_{m,h},\widehat{c}^{*}_{m,h},p^{*}_{m,h}\}$, the associated solution to $\mathsf{P}$ can be recovered as $\{d^{*}_{m,h}=\mathrm{exp}(\widehat{d}^{*}_{m,h}),c^{*}_{m,h}=\mathrm{exp}({\widehat{c}^{*}_{m,h}}),p^{*}_{m,h}\}$. 
Since $\mathsf{P}_1$ involves $3MH$ variables, the total complexity of solving $\mathsf{P}_1$ with GP is $\mathcal{O}((3MH)^{3.5})$. 

\begin{figure}[!t]
    \centering
    \includegraphics[width=0.48\textwidth]{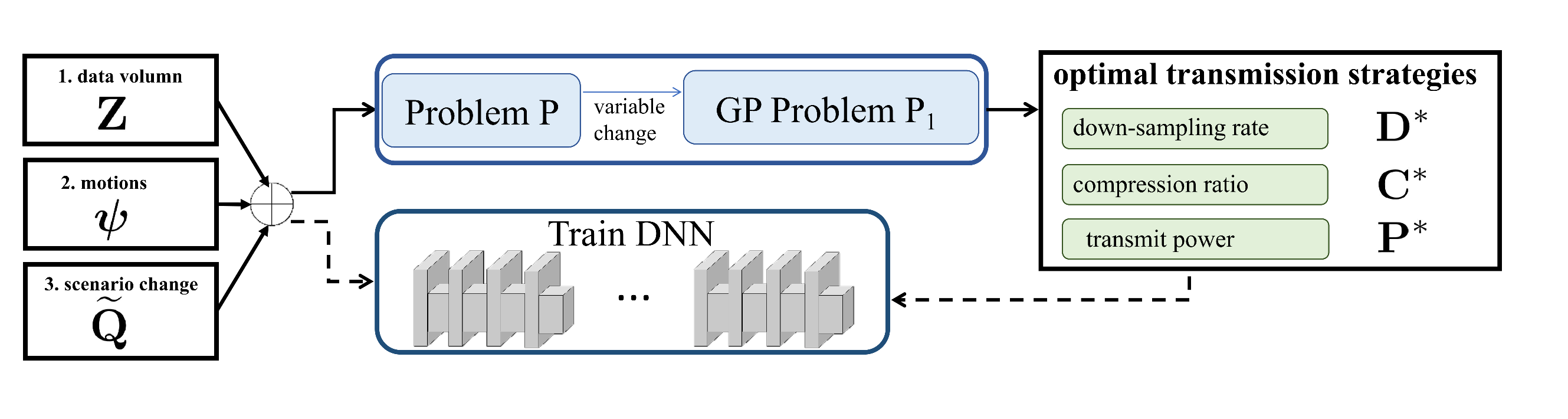} 
    \caption{Pipeline of the LTO-IPMC algorithm.}
    \label{fig:LTO_schems} 
\end{figure}

\subsection{Learning to Optimize}

In Section IV-A, we have developed a GP-based method to solve the original problem. However, as mentioned below $\mathsf{P}_1$, the complexity of the GP algorithm is still high when the number of sub-durations $H$ is large, which makes it impractical for real-time implementation. 

To address this challenge, we propose an LTO-based method for faster IPMC design. 
In particular, LTO trains an agent to accomplish tasks by learning from experts' demonstrations. The core advantage of LTO is to convert time-consuming iterative computations into real-time feed-forward inference. 
In this work, we adopt the demonstration dataset generated by the GP-based algorithm. 
Then, we use these demonstrations to train our DNN model in an offline manner using back propagation. Afterward, the trained DNN model is deployed at the robot to conduct real-time inference by imitating the GP solver. The entire architecture of LTO is shown in Fig.~\ref{fig:LTO_schems}, which consists of three phases:

 \emph{1) Offline Training Sample Generation}: We leverage the GP algorithm to generate the demonstration dataset 
$\mathcal{D} = \{\mathcal{S}_i \}_{i=1}^I$, where the $i$-th sample is given by
\begin{align}
    \mathcal{S}_i  = \{(\mathbf{Z}^{(i)},\widetilde{\mathbf{Q}}^{(i)},\bm{\psi}^{(i)}),(\mathbf{D}^{*(i)},\mathbf{C}^{*(i)},\mathbf{P}^{*(i)})\}.
\end{align}
The input tuple $(\mathbf{Z}^{(i)},\widetilde{\mathbf{Q}}^{(i)},\bm{\psi}^{(i)})$ is generated by collecting multi-modal sensor data using the robot. 
The output tuple $(\mathbf{D}^{*(i)},\mathbf{C}^{*(i)},\mathbf{P}^{*(i)})$ is generated by the GP-based algorithm. 

\emph{2) Offline DNN Model Training}: We employ a five-layer fully connected network with progressively increasing hidden layer widths (64, 128, 256, 512)\cite{LTO2} to learn the mapping from the input to the output tuple. The output layer has 6 units, equal to the dimensionality of the variable space. This five-layer FC architecture was selected as the optimal balance, achieving high accuracy while maintaining a compact model size ($<$1 MB) and low inference latency.
     The first four layers utilize Rectified Linear Unit (ReLU) activation functions. For model training, we use the Mean Squared Error (MSE) as the loss function and optimize the network parameters using the Adam (Adaptive Moment Estimation) optimizer.

\emph{3) Online Decision Making}: With the pre-trained DNN, robot can obtain real-time decisions (including down-sampling ratios, compression ratios, transmit powers) by estimating the data volume matrix $\mathbf{Z}$, normalized scenario change-rate matrix $\widetilde{\mathbf{Q}}$, motion-aware vector $\bm{\psi}$, and feeding them into the DNN.

 \begin{figure}[!t]
     \centering 
     \begin{subfigure}[b]{0.24\textwidth}
         \includegraphics[width=\linewidth]{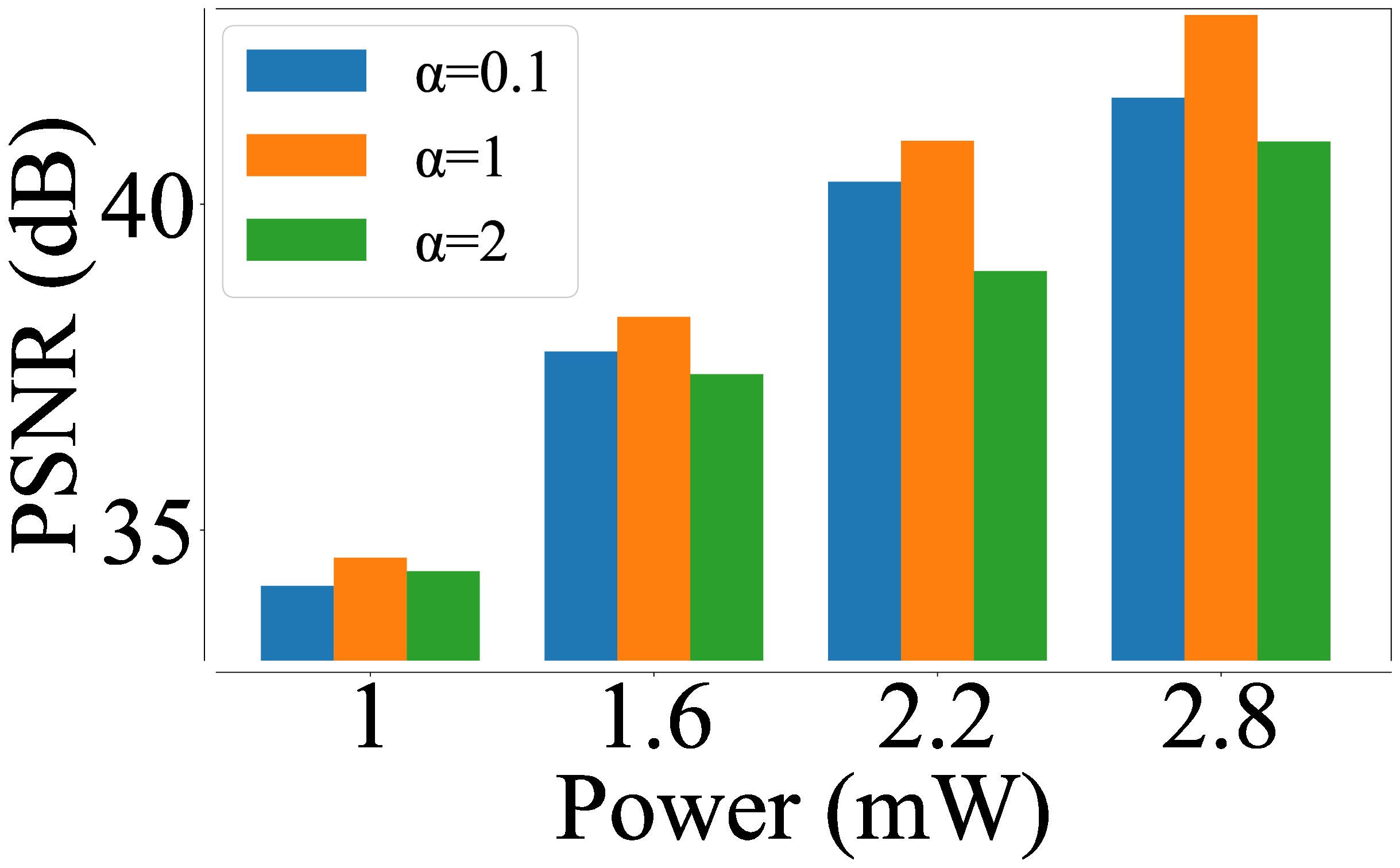}
         \caption{PSNR vs $P$ under various $\alpha$.}
         \label{fig:tune_psnr}
     \end{subfigure}
     \begin{subfigure}[b]{0.24\textwidth}
         \includegraphics[width=\linewidth]{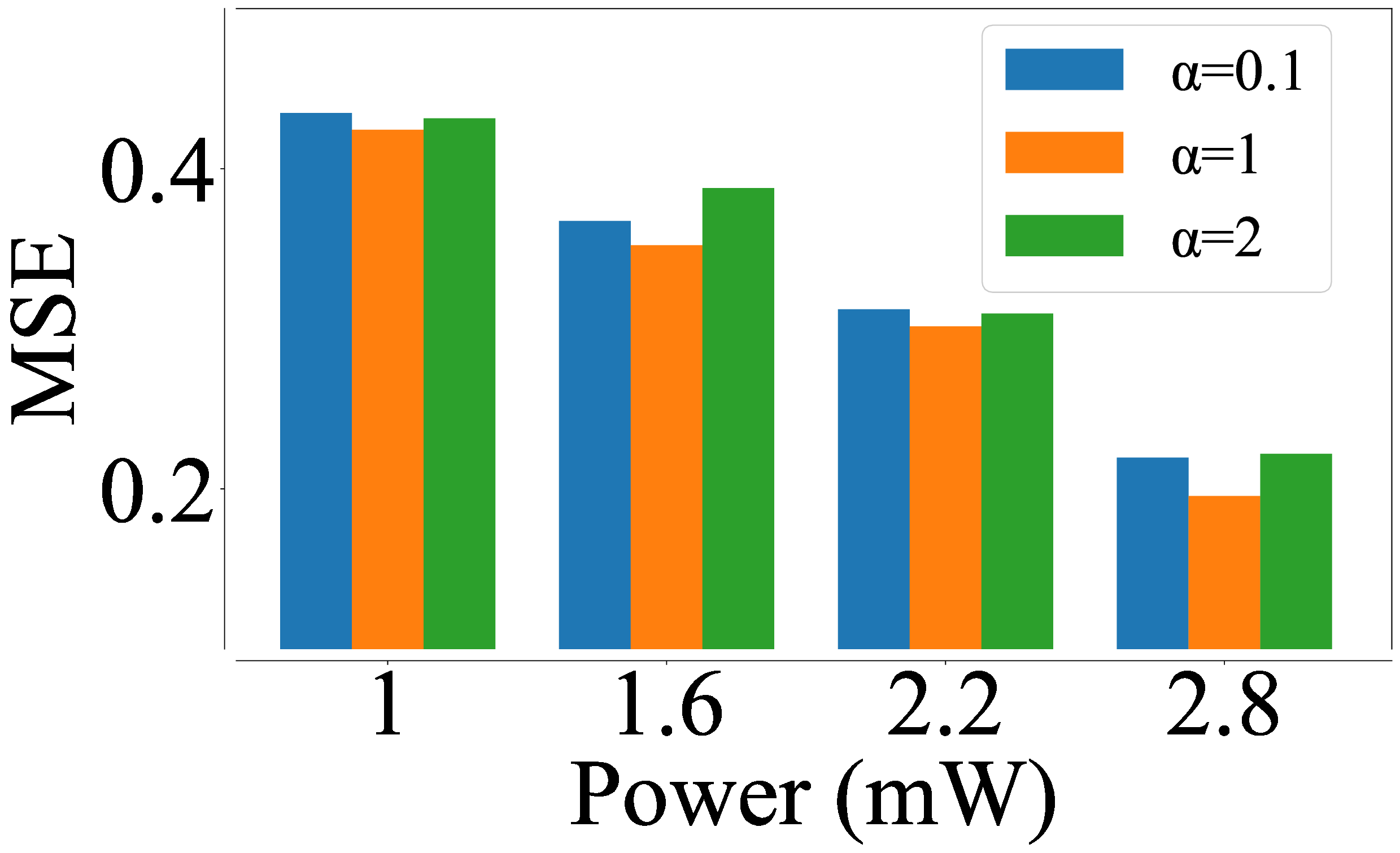}
         \caption{MSE vs $P$ under various $\alpha$.}
         \label{fig:tune_mse}
     \end{subfigure}
     \\ 
     \begin{subfigure}[b]{0.24\textwidth}
         \includegraphics[width=\linewidth]{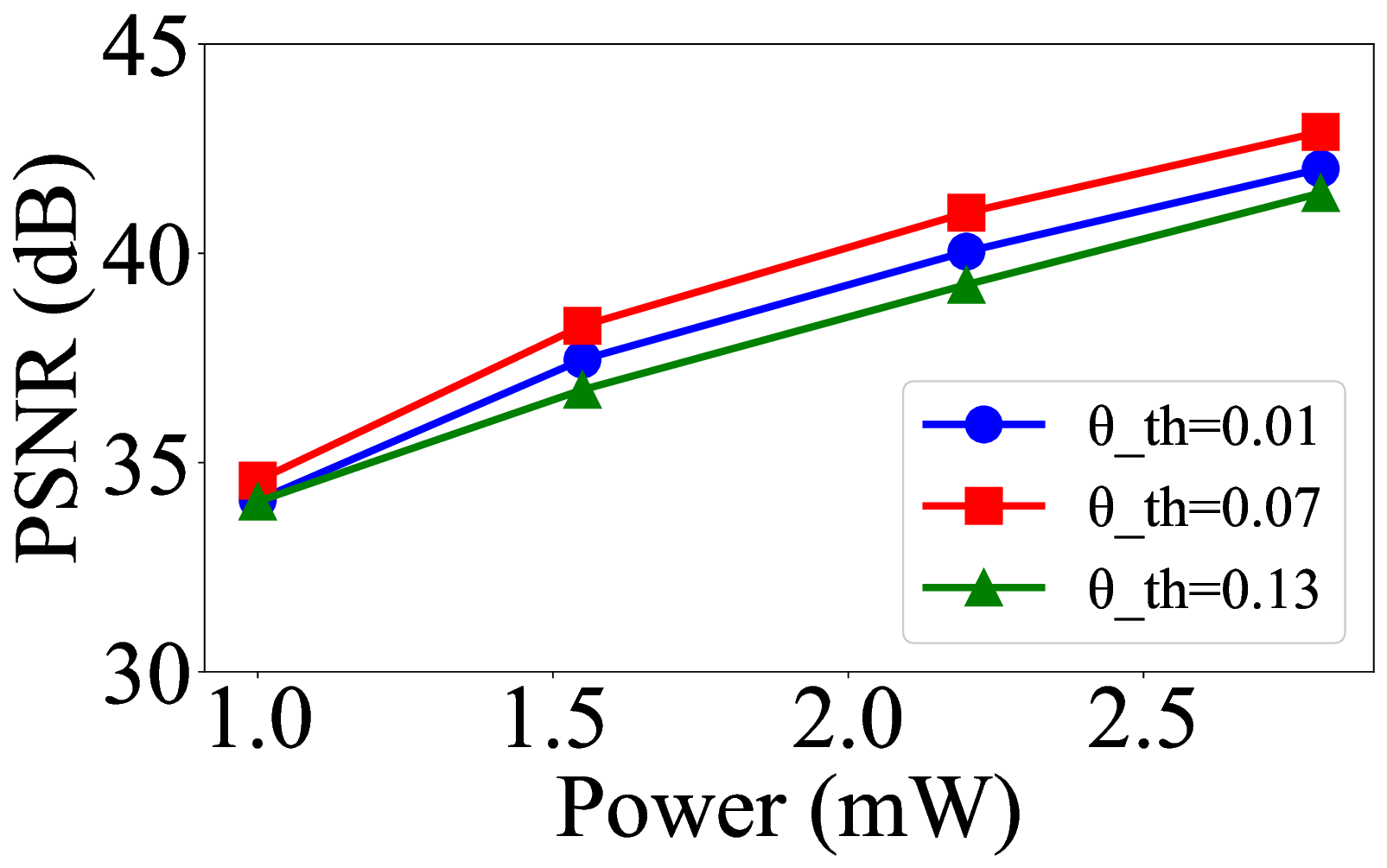}
         \caption{{PSNR vs $P$ under various $\theta_{\mathrm{th}}$.}}
         \label{fig:theta_tune_psnr}
     \end{subfigure}
     \begin{subfigure}[b]{0.24\textwidth}
         \includegraphics[width=\linewidth]{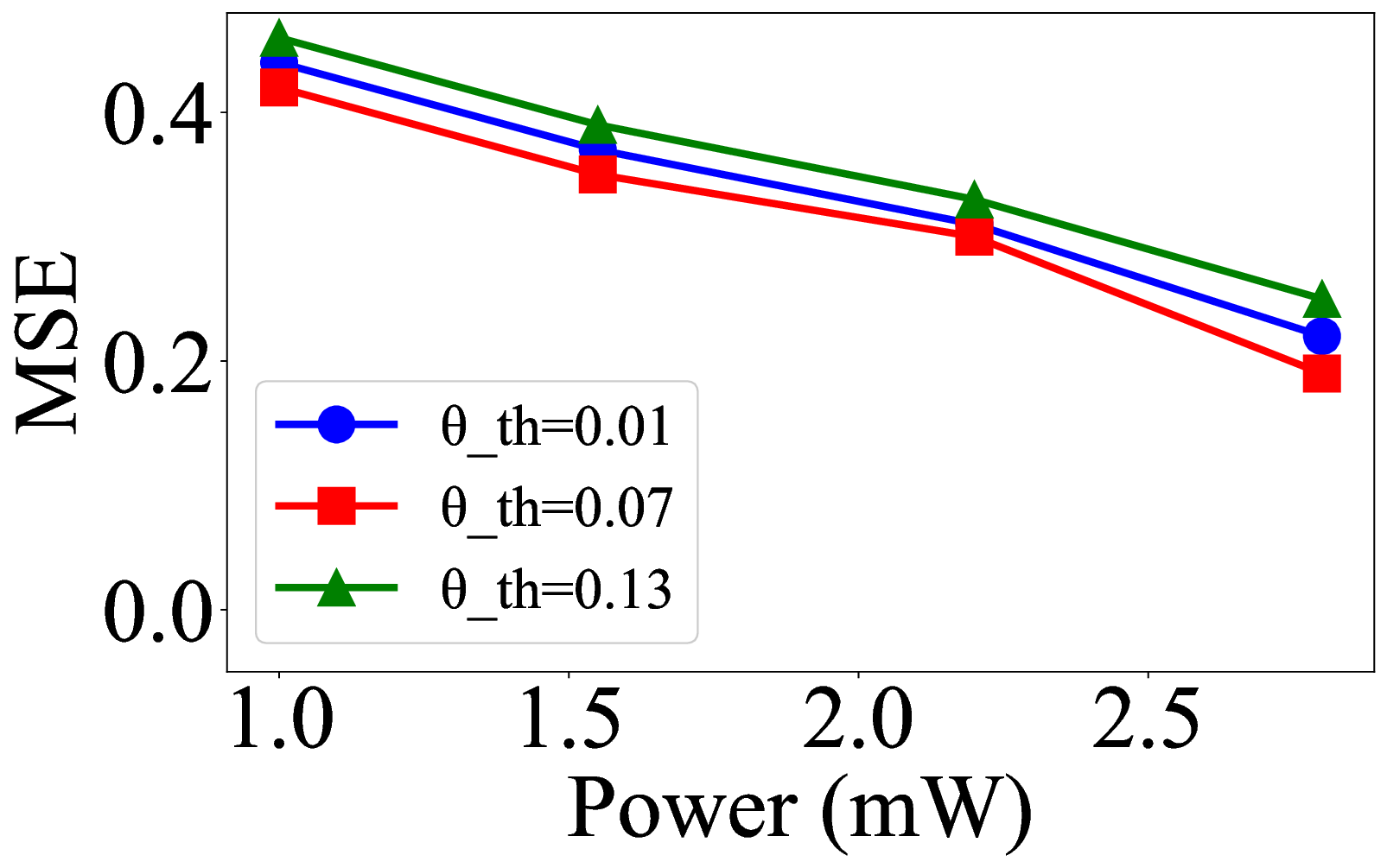}
         \caption{{MSE vs $P$ under various $\theta_{\mathrm{th}}$.}}
         \label{fig:theta_tune_mse}
     \end{subfigure}
       \\
     \begin{subfigure}[b]{0.24\textwidth}
         \includegraphics[width=\linewidth]{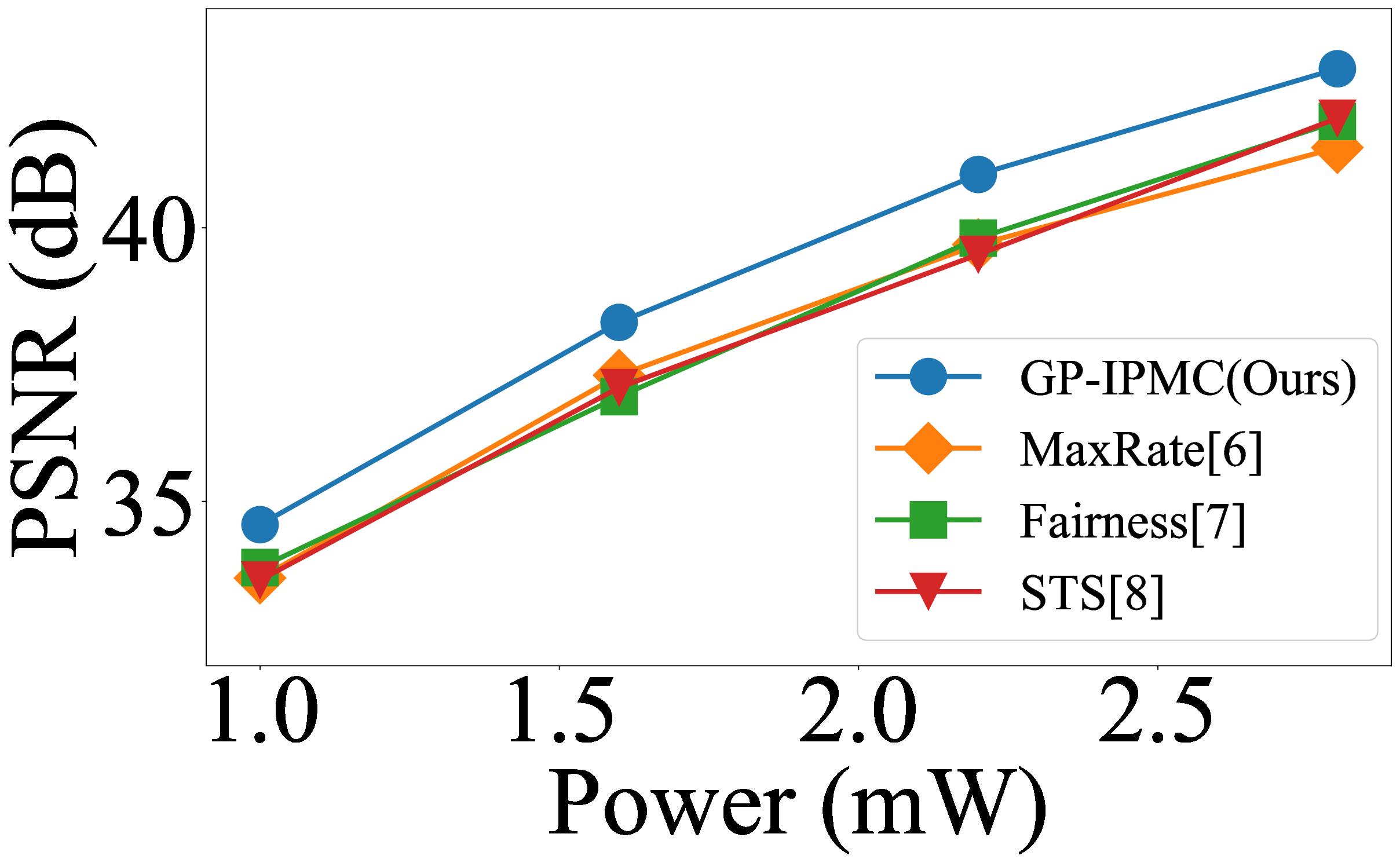}
         \caption{Comparison of PSNRs.}
         \label{fig:psnr}
     \end{subfigure}
     \begin{subfigure}[b]{0.24\textwidth}
         \includegraphics[width=\linewidth]{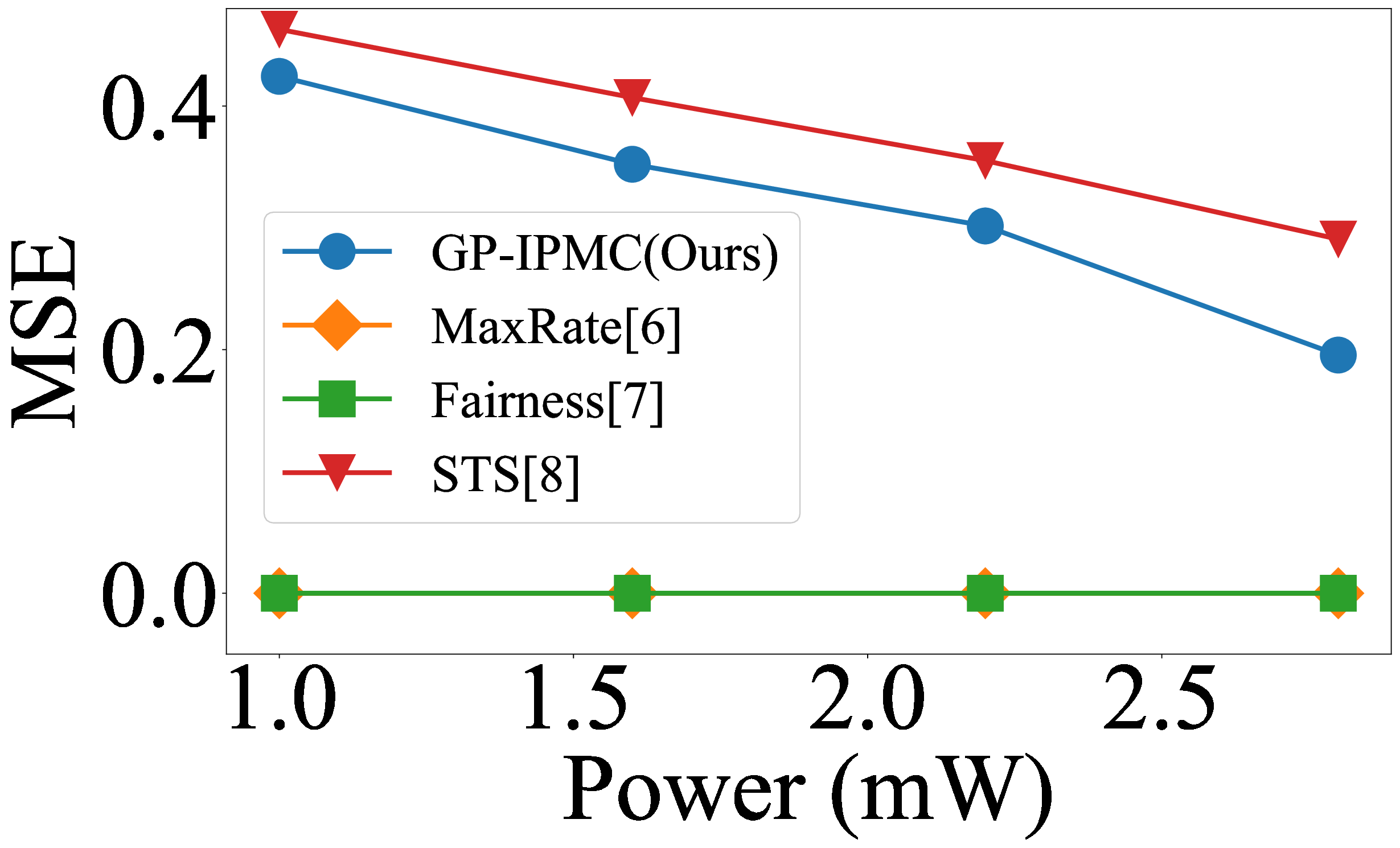}
         \caption{Comparison of MSEs.}
         \label{fig:mse}
     \end{subfigure}
     \caption{Quantitative comparison of PSNRs and MSEs.}
     \label{fig:IPMC}
     \vspace{-0.1in}
 \end{figure}

\begin{figure}[!t]
    \centering
    \includegraphics[width=0.5\textwidth]{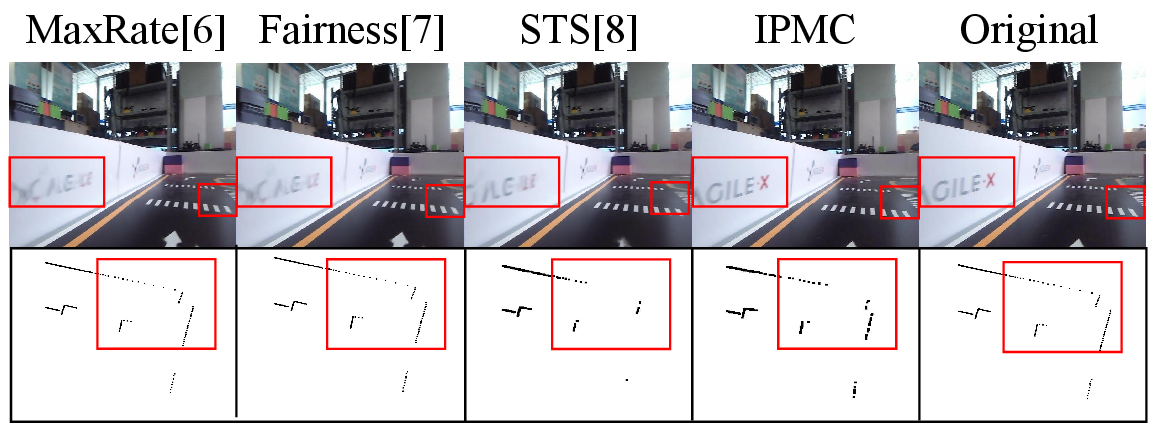} 
    \caption{Visualization of reconstructed camera and lidar data.}
    \label{fig:recover} 
    \vspace{-0.15in}
\end{figure}

\begin{figure*}[!t]
    \centering
    \begin{minipage}[b]{\textwidth}
        \centering       
        \begin{minipage}[b]{0.19\textwidth}
            \includegraphics[width=\textwidth]{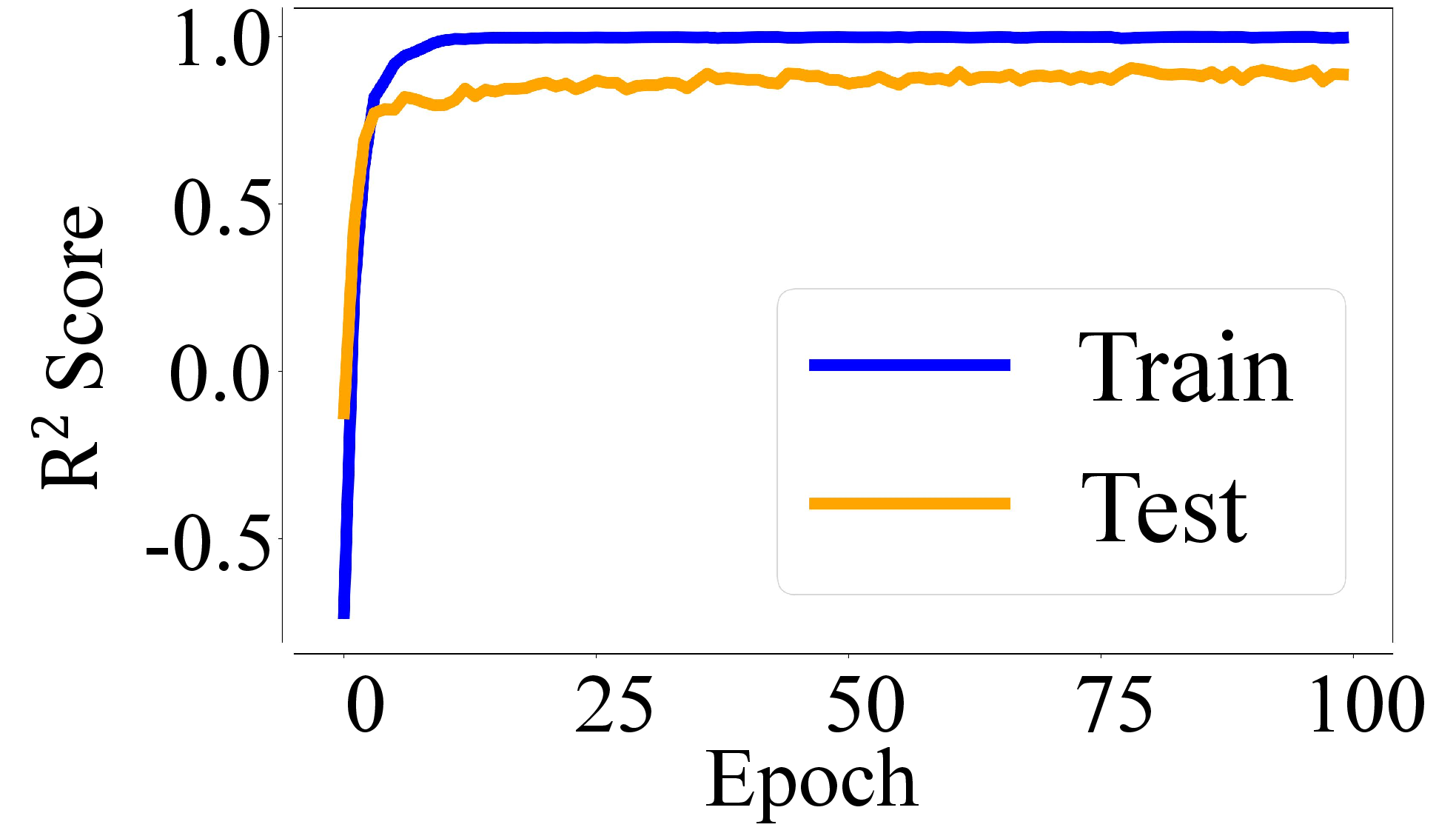}
            \vspace{-0.8\baselineskip}
            \centerline{\footnotesize (a)}
        \end{minipage}
        \hfill
        \begin{minipage}[b]{0.19\textwidth}
            \includegraphics[width=\textwidth]{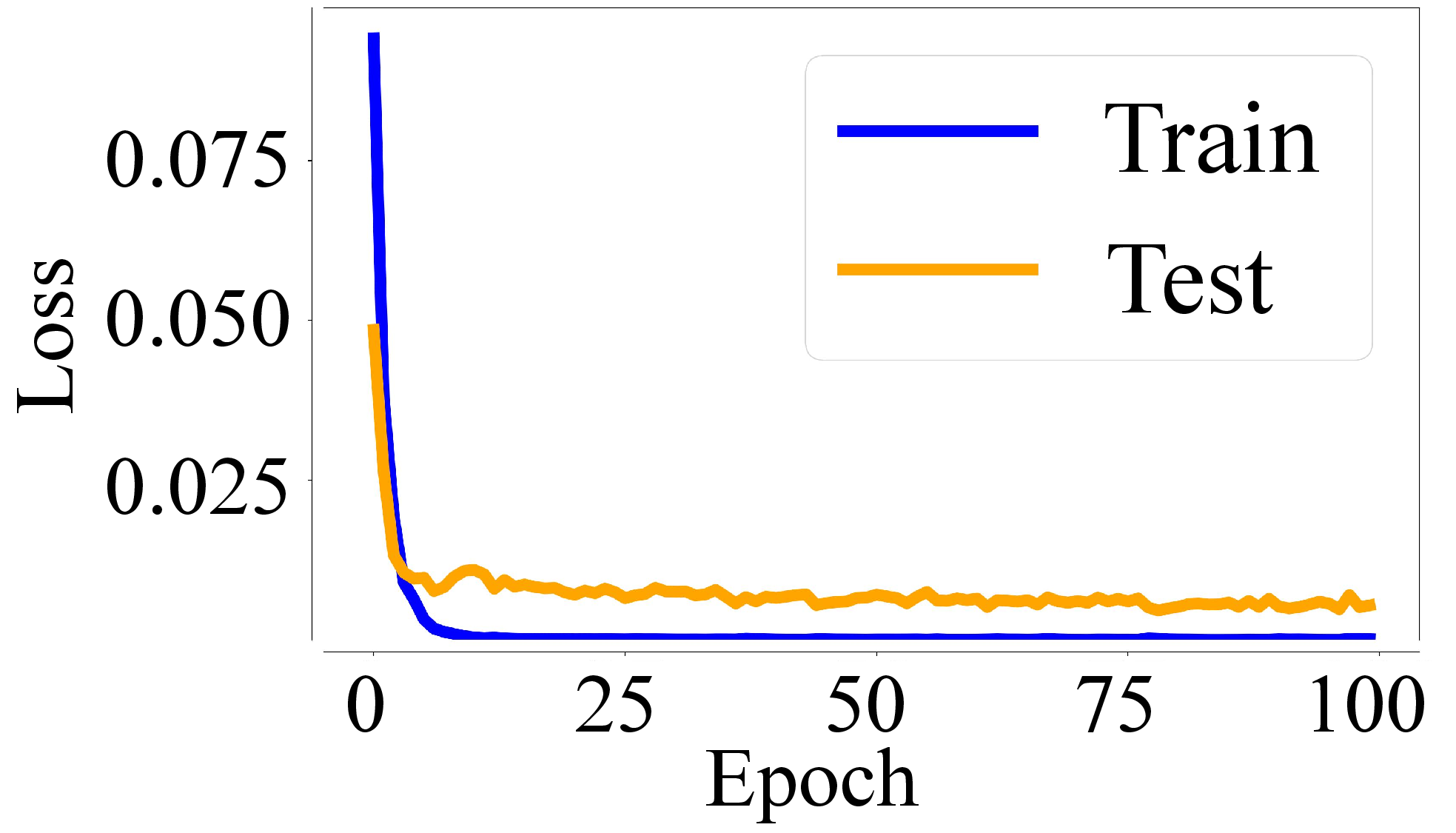}
            \vspace{-0.8\baselineskip}
            \centerline{\footnotesize (b)}
        \end{minipage}
        \hfill
        \begin{minipage}[b]{0.19\textwidth}
            \includegraphics[width=\textwidth]{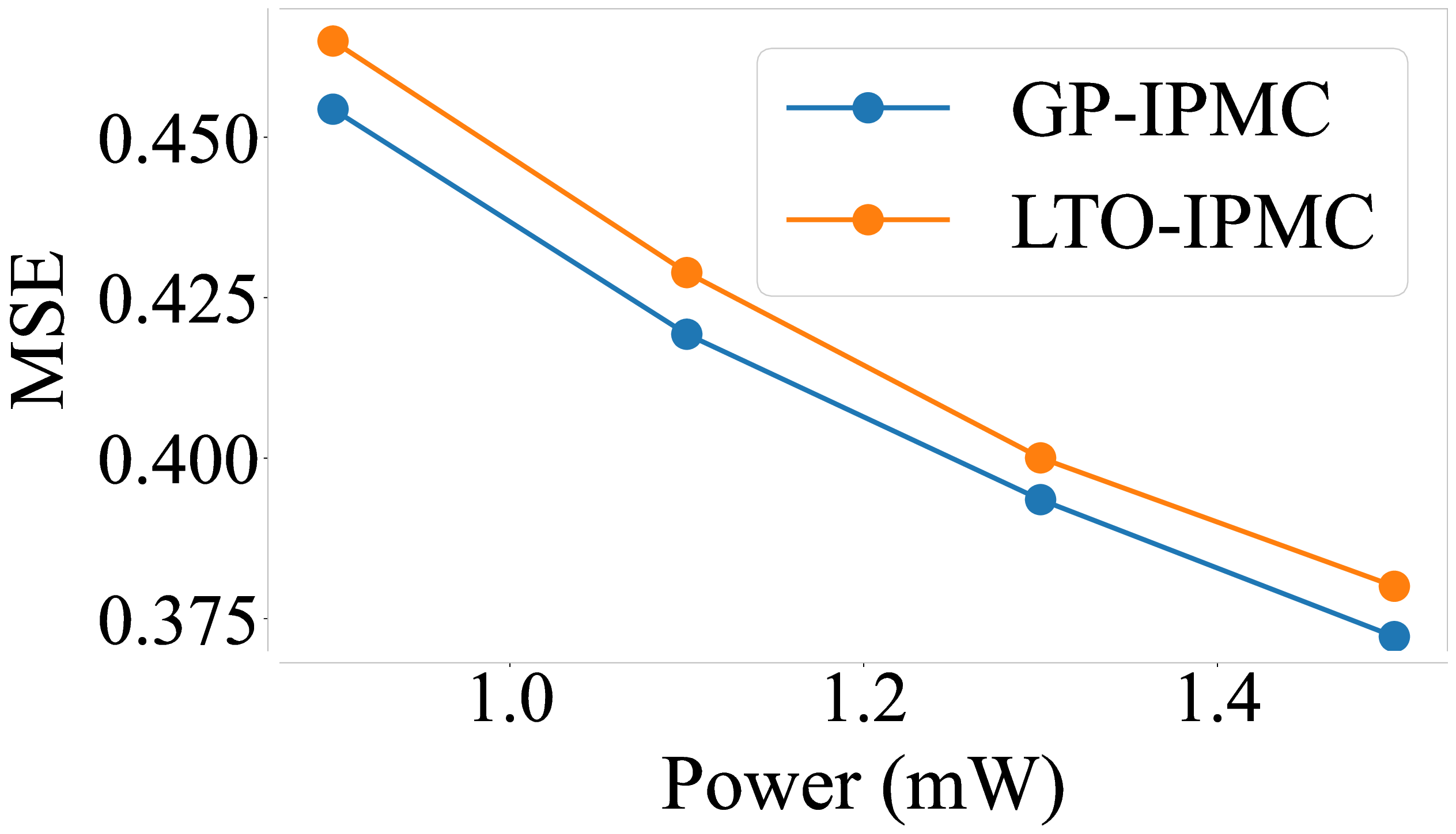}
            \vspace{-0.8\baselineskip}
            \centerline{\footnotesize (c)}
        \end{minipage} 
        \hfill
        \begin{minipage}[b]{0.19\textwidth}
            \includegraphics[width=\textwidth]{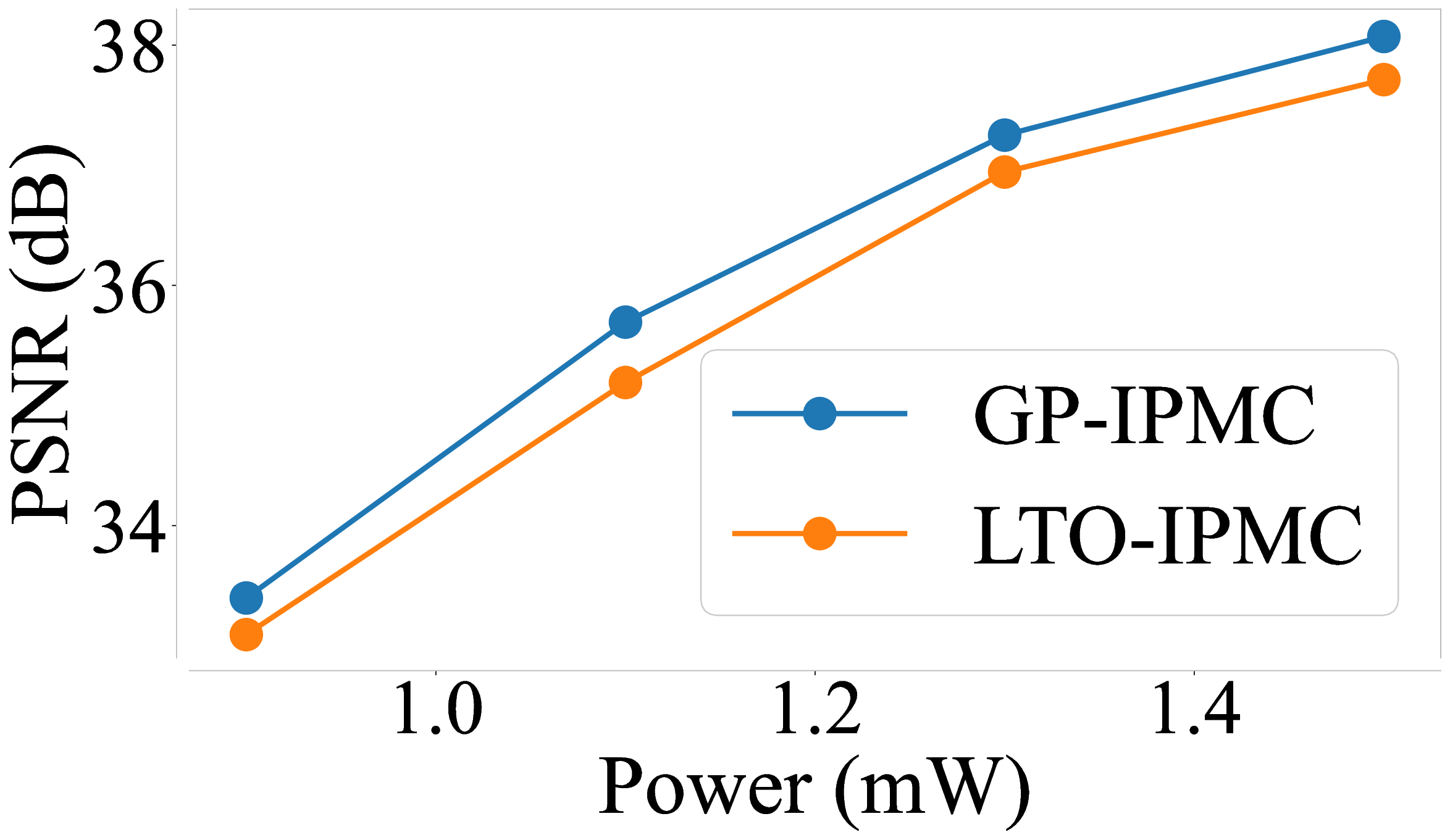}
            \vspace{-0.8\baselineskip}
            \centerline{\footnotesize (d)}
        \end{minipage}
        \hfill
        \begin{minipage}[b]{0.19\textwidth}
            \includegraphics[width=\textwidth]{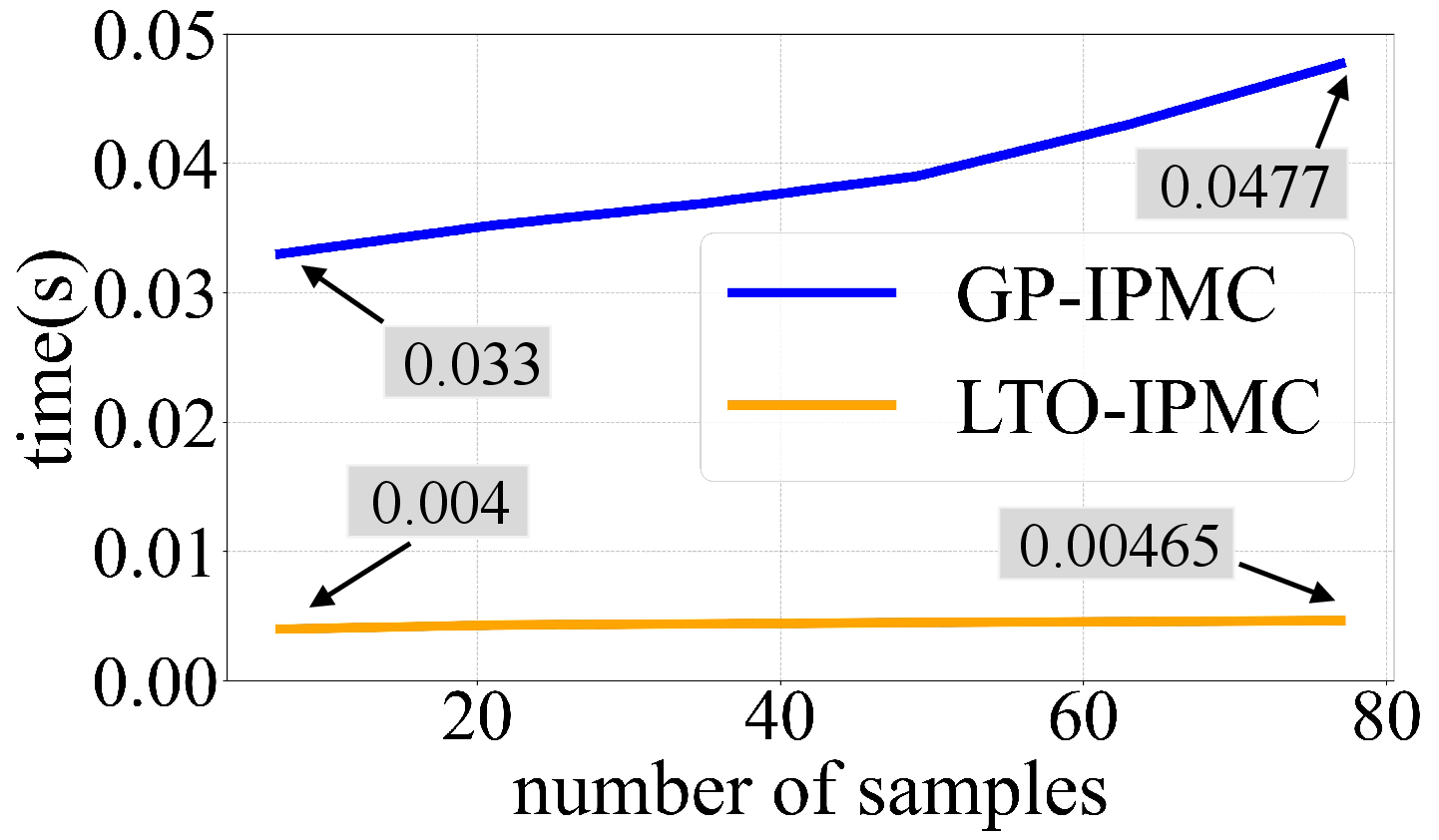}
            \vspace{-0.8\baselineskip}
            \centerline{\footnotesize (e)}
        \end{minipage}
        \vspace{0.2cm}
        \caption{a) Training and test accuracies of LTO; b) Training and test losses of LTO; c) MSE versus $P$ for GP-IPMC and LTO-IPMC; d) PSNR versus $P$ for GP-IPMC and LTO-IPMC; e) Comparison of computation time.}
        \label{fig:LTO}
             \vspace{-0.05in}
    \end{minipage}
\end{figure*}

\vspace{-0.05in}

\section{Simulation Results}\label{section5}

This section presents simulation results to evaluate the performance of the proposed IPMC. Experiments are conducted on a Ubuntu workstation with $3.7$ GHz CPU and Nvidia $3050$ GPU. 
The dataset is collected using the LIMO robot  for a total duration of $T=10137$\,s. 
The entire duration is divided into $H=10137$ sub-durations and each sub-duration lasts $1$ second. 
Each sub-duration is further divided into $L=7$ frames with $\Delta t=0.1428$\,s, corresponding to a sensing frequency of $7$\,Hz.
The communication bandwidth $B=1\,\text{MHz}$ and noise power $\sigma=-80\,\text{dBm}$. 
The channel gain $G_{m,h}\sim \mathcal{U}(-80,-85)$\,dB, where $\mathcal{U}(a,b)$ denotes uniform distribution within interval $[a,b]$.
The minimum sampling and compression ratios are set to  $d_{\text{min}}=0.1$ and $c_{\text{min}}= 0.8$.  

We simulate the case of $M=2$.
The camera image is denoted as $\mathbf{x}_{1,h,l}\in {\mathbb{R}^{N_1 \times 1}}$, 
where $N_1=3IW$. 
Here, $I$ and $W$ are the length and width of camera images respectively, and coefficient $3$ accounts for the RGB channels.
The lidar point cloud data is denoted as $\mathbf{x}_{2,h,l}\in {\mathbb{R}^{N_{2} \times 1}}$, where $N_{2}=2O$, $O$ is the number of points, and coefficient $2$ accounts for the points coordinates.
The data volumes of each data are $311352$ and $71168$ bits for camera and lidar, respectively. The bounds for the control vector are defined as $\mathbf{u}_{\min}$= [$-1$\,m/s, $-1$\,rad/s] and $\mathbf{u}_{\max}$= [$1$\,m/s, $1$\,rad/s]. The acceleration vector is subject to the constraints $\bm{\beta}_{\min}$=  [$-0.1$\,m/s, $-0.05$\,rad/s] and $\bm{\beta}_{\max}$=  [$0.1$\,m/s, $0.05$\,rad/s].

First, we conduct experiments to evaluate the performance of GP-IPMC under different $\alpha$.
It can be seen from Fig.~\ref{fig:IPMC}a and Fig.~\ref{fig:IPMC}b that as $\alpha$ increases, the PSNR (MSE) first increases (decreases) then decreases (increases), meaning that the optimal choice of $\alpha$ must be finite and lie within the interval $[0.1,2]$.
Moreover, in our experiment, $\alpha$=1 yields the best performance, and we select $\alpha$=1 in the subsequent part.

Then, we evaluate the GP-IPMC under different values of $\theta_{\mathrm{th}}\in\{0.01, 0.07, 0.13\}$.
The result is shown in Fig. 3c and Fig. 3d.
It can be seen that with the choice of $\theta_{\mathrm{th}}=0.07$, the PSNR achieves its highest value and the MSE achieves its lowest value.
This means that the best transmission performance is obtained at $\theta_{\mathrm{th}}=0.07$, which corroborates Fig.~\ref{fig:fig1}c. 
This also confirms that $\theta_{\mathrm{th}}$ indeed has a non-negligible impact on the edge robotic system.

Next, we compare the GP-IPMC to the following baselines: 
1) MaxRate\cite{sumrate_maximization}, which maximizes the sum-rate of ER; 2) Fairness\cite{maxmin_fairness}, which maximizes the minimum frame-rate of ER; 
and 3) STS\cite{STS}.
It can be seen from Fig. \ref{fig:IPMC}e that the proposed GP-IPMC achieves the highest PSNR value among all the simulated scheme and improves PSNR by at least $4.03\%$ compared to the second-best method. 
As for the lidar data transmission, the GP-IPMC scheme also achieves satisfactory MSE performance, as shown in Fig. \ref{fig:IPMC}f.

Note that the MaxRate and Fairness schemes ignore the multi-modality balancing, allocating excessive resources to the lidar data. 
This can also be observed from Fig.~\ref{fig:recover}, where the MaxRate and Fairness schemes lead to the most blurred images.
Our GP-IPMC method effectively mitigates this modality issue by adopting the robot-oriented cost $C_0$. 
Consequently, the image and point-cloud data of GP-IPMC are both identical to the original ground truth data.
We also observe that the STS scheme improves the image quality compared to MaxRate and Fairness schemes. 
This is because the STS scheme incorporates modality balancing and adaptive transmission mechanisms.
However, it degrades the lidar transmission performance (missing part of points as marked in red box). 
Moreover, there exist significant discrepancies of caption ``AGILEX'' on the board between STS and GP-IPMC. 

Finally, we evaluate the proposed LTO-IPMC in Section\ref{section4}. 
We take $10000$ samples as the training dataset, and the remaining $137$ samples as test dataset.
We train the DNN for $100$ epochs with a learning rate of $0.01$ and a batch size of $1024$. We use R-squared score to measure the accuracy of DNN. 
Fig.~\ref{fig:LTO}a and Fig.~\ref{fig:LTO}b illustrate the training and testing accuracies and losses, respectively.
It can be seen that the proposed LTO-IPMC achieves an accuracy of $0.98$ and a loss of $0.003$, meaning that LTO successfully learns how to imitate GP. 
The MSE and PSNR performances of GP-IPMC and LTO-IPMC are compared in Fig.~\ref{fig:LTO}c and Fig.~\ref{fig:LTO}d.
It is found that the image and point-cloud qualities of LTO-IPMC are slightly worse than those of GP-IPMC. 
However, such a degradation is acceptable, given the fact that LTO-IPMC works significantly faster than GP-IPMC, as shown in Fig.~\ref{fig:LTO}e.
The computation time of LTO is less than $10\,\%$ compared to that of GP, and the speed improvement of LTO is more significant as the number of samples increases.
Particularly, the execution time of LTO-IPMC is only $4$ milliseconds, which demonstrates the real-time execution capability of LTO-IPMC.

\vspace{-0.05in}

\section{Conclusion}\label{section6}

This paper presents the IPMC framework, which enables real-time robot-oriented transmission for power-constrained ER systems, by jointly optimizing perception, motion, and communication functionalities.
Experiments show that the proposed GP-IPMC outperforms existing benchmarks by $4.03\%$ and $8.69\%$, in terms of PSNR and MSE, respectively. 
Moreover, the proposed LTO-IPMC approach achieves $10$× faster computation than GP-IPMC while ensuring close-to-optimal performance. 
It is also found that the IPMC achieves an excellent balance between different data modalities.

\bibliographystyle{IEEEtran}
\bibliography{main}
\end{document}